\documentclass{article}

\usepackage[preprint]{neurips_2026}

\usepackage[utf8]{inputenc} 
\usepackage[T1]{fontenc}    
\usepackage{url}            
\usepackage{booktabs}       
\usepackage{amsfonts}       
\usepackage{nicefrac}       
\usepackage{microtype}      
\usepackage{xcolor}         
\usepackage{amsmath}
\usepackage{subcaption}
\usepackage{graphicx}
\usepackage{multirow}
\usepackage{algorithm}
\usepackage{algpseudocode}
\usepackage{hyperref}       

\title{LPNSR: Learnable Noise Prediction for Diffusion-Based Image Super-Resolution}

\author{%
  Shuwei Huang, Shizhuo Liu, Zijun Wei\\
  Huazhong University of Science and Technology \\
  \texttt{\{frozen2001, shizhuol\}@hust.edu.cn}, \texttt{weiiizong1001@gmail.com}\\
}

\begin{document}

\maketitle

\begin{abstract}
  Diffusion-based image super-resolution (SR) aims to reconstruct high-resolution (HR) images from low-resolution (LR) observations.
  A key property of diffusion models is that, once the starting point of the reverse chain and the denoising network are fixed, the quality of the generated image is dictated by the noise maps sampled at the intermediate steps, which are drawn from an unconstrained standard Gaussian distribution in conventional pipelines.
  This property raises a natural question: does there exist a noise sampling distribution better than the standard Gaussian that improves the quality of the generated images?
  To this end, we use a parameterized deep neural network to predict the mean and the variance of the Gaussian noise sampling distribution at each intermediate step, and design two training schemes, one supervised by the quality of the final generated image and the other by aligning each reverse step with the forward posterior.
  Experiments show that, compared with the standard Gaussian distribution, the learnable noise sampling distribution improves the quality of the generated images.
  The source code of our method can be found at \url{https://github.com/Faze-Hsw/LPNSR}.
\end{abstract}

\section{Introduction}
\label{sec:Introduction}
Image super-resolution (SR) aims to reconstruct a high-resolution (HR) image from its low-resolution (LR) observation, a severely ill-posed problem since the degradation is unknown and multiple HR images are consistent with the same LR input.
Since SRCNN \citep{dong2015image} first formulated SR as regression with a deep network, learning-based methods have dominated this task.
Early regression-based methods \citep{kim2016accurate,wang2015deep,ahn2018image} steadily improved reconstruction fidelity, yet they inherently yield over-smoothed results, since pixel-wise objectives push the prediction toward the conditional mean over plausible HR images \citep{blau2018perception}.
GAN-based methods \citep{ledig2017photo,sajjadi2017enhancenet,menon2020pulse} instead recover realistic textures with adversarial and perceptual losses, and their successors improve the generator design and the degradation modeling for real-world inputs \citep{wang2018esrgan,zhang2021designing,wang2021real}, but adversarial training remains unstable and may hallucinate unfaithful textures.
More recently, diffusion models \citep{NEURIPS2020_4c5bcfec,song2020denoising,Rombach_2022_CVPR} have become the dominant paradigm for SR, generating the HR image by an iterative denoising process that is conditioned on the LR input \citep{9887996} or guided by it during sampling \citep{choi2021ilvr,Chung_2022_CVPR,NEURIPS2022_95504595,chung2022diffusion}.

A key property of diffusion models is that, once the starting point of the reverse chain and the denoising network are fixed, the noise maps sampled at the intermediate steps are the only degree of freedom of the generated image, and the quality of the generated image is dictated by them (see Fig.~\ref{fig:inference_steps_comparison}).
In conventional pipelines, however, these noise maps are drawn from a fixed standard Gaussian distribution.
This raises a natural question: does there exist a noise sampling distribution better than the standard Gaussian that improves the quality of the generated images?

A closely related idea is diffusion inversion, which seeks the noise maps with which a pretrained diffusion model reconstructs a given target image, and has been developed mainly for image editing and personalization.
Early works optimize the conditioning signal used by the sampler, such as the textual or visual embedding \citep{gal2022image,mokady2023null,miyake2025negative,nguyen2023visual}, whereas later works operate directly in the noise space, correcting the approximation error of the inversion through coupled transformations, fixed-point iterations, or re-designed injection schedules \citep{wallace2023edict,meiri2023fixed,ju2023direct,kang2024eta}.
In parallel, diffusion models are widely used as priors for inverse problems, where the intermediate latents are guided toward the observation during sampling \citep{choi2021ilvr,Chung_2022_CVPR,NEURIPS2022_95504595,chung2022diffusion,fei2023generative,song2023pseudoinverse}.
Recently, this idea has also been extended to SR: InvSR \citep{Yue_2025_CVPR} trains a network to predict a noise map that serves as the starting point of a frozen text-to-image diffusion model.

In contrast to these methods, which determine the injected noise either by per-input optimization or by predicting a single noise map that only serves as the starting point of a frozen model, we learn the noise sampling distribution at each intermediate step as a conditional distribution, which can be evaluated in a single forward pass.
Specifically, we parameterize the noise injected at each intermediate step as a diagonal Gaussian, whose mean and variance are predicted by a deep neural network conditioned on the current latent, the clean-image estimate of the frozen denoiser, the low-resolution latent, and the time step.
We then design two training schemes for this noise predictor.
The first one is an end-to-end supervised training: the predictor is plugged into every intermediate step of the reverse chain, the whole chain is unrolled, and the predictor is optimized by supervising the final generated image with pixel-wise, perceptual \citep{zhang2018unreasonable}, and adversarial \citep{goodfellow2014generative} losses.
The second one trains the predictor step by step without unrolling the chain: based on the alignment between the model reverse distribution and the forward posterior, the network is trained to regress the prediction residual of the frozen denoiser, i.e., the gap between the ground-truth clean latent and its denoised estimate, which amounts to adding a mean-correction term to the reverse distribution.
Both schemes leave the pretrained denoising network and the original inference mechanism untouched.

The main contributions of this work are summarized as follows.
First, we model the injected noise as a learnable conditional distribution, and propose two training schemes, i.e., the end-to-end supervised training and the mean-correction training.
Second, we propose LPNSR (Learnable Prediction of Noise for Super-Resolution) on the residual-shifting diffusion paradigm \citep{NEURIPS2023_2ac2eac5}, where a multi-input-aware noise predictor estimates the mean and the variance of the injected noise at each sampling step, while the pretrained denoising network and the inference mechanism remain untouched.
Third, we conduct experiments and show that the learnable noise sampling distribution improves the quality of the generated images over the standard Gaussian distribution.

\section{Related work}
\label{sec:Related Work}
\textbf{Image Super-Resolution.} Along with the proliferation of deep learning, deep learning-driven approaches have progressively emerged as the dominant paradigm for SR \citep{dong2015image,wang2021srsurvey}.
Early prominent works primarily focused on training regression models using paired LR-HR data \citep{ahn2018image,kim2016accurate,wang2015deep}. Though these models effectively capture the expectation of the posterior distribution, they inherently suffer from over-smoothing artifacts in generated results \citep{ledig2017photo,menon2020pulse,sajjadi2017enhancenet}.
To enhance the perceptual quality of reconstructed HR images, generative SR models have garnered growing interest, including autoregressive architectures \citep{dahl2017pixel,menick2018generating,van2016conditional,parmar2018image}. Despite notable gains in perceptual performance, autoregressive models typically incur substantial computational overhead.
Additionally, GAN-based SR methods have attained remarkable success in perceptual quality \citep{ledig2017photo,sajjadi2017enhancenet,menon2020pulse,wang2018esrgan,wang2021real}, yet the training process remains unstable.
More recently, diffusion-based models have become a focal point of SR research \citep{choi2021ilvr,Chung_2022_CVPR,NEURIPS2022_95504595,Rombach_2022_CVPR,9887996}. These methods generally fall into two categories: those that concatenate the LR image to the denoiser's input \citep{Rombach_2022_CVPR,9887996}, and those that adapt the reverse process of a pretrained diffusion model \citep{choi2021ilvr,Chung_2022_CVPR,NEURIPS2022_95504595}. 
While these diffusion-based approaches yield promising performance, they still introduce unconstrained random Gaussian noise in each step of the reverse diffusion process, rather than meaningful noise maps.

\textbf{Diffusion Inversion.} This paradigm aims to find the noise maps with which a pretrained diffusion model reconstructs a given target image, and has been developed mainly for image editing and personalization. Early works optimized text embeddings for better alignment \citep{gal2022image,mokady2023null}, and follow-up works further refined inversion via textual/visual prompts \citep{miyake2025negative,nguyen2023visual} or intermediate noise map optimization \citep{ju2023direct,kang2024eta,meiri2023fixed,wallace2023edict}. 
However, existing methods determine the noise either by per-input optimization or by predicting a single noise map per input, and none of them attempts to formulate the noise sampling distribution at each intermediate step as a parametric distribution.
For SR, InvSR \citep{Yue_2025_CVPR} trains a noise predictor that produces the starting point of a frozen text-to-image diffusion model, while the noise maps injected at the intermediate steps are still drawn from a standard Gaussian distribution, as in conventional pipelines.

\section{Methodology}
\label{sec:Methodology}

\subsection{Motivation}
\label{sec:Motivation}

\begin{figure}
\centering
\includegraphics[width=\linewidth, keepaspectratio]{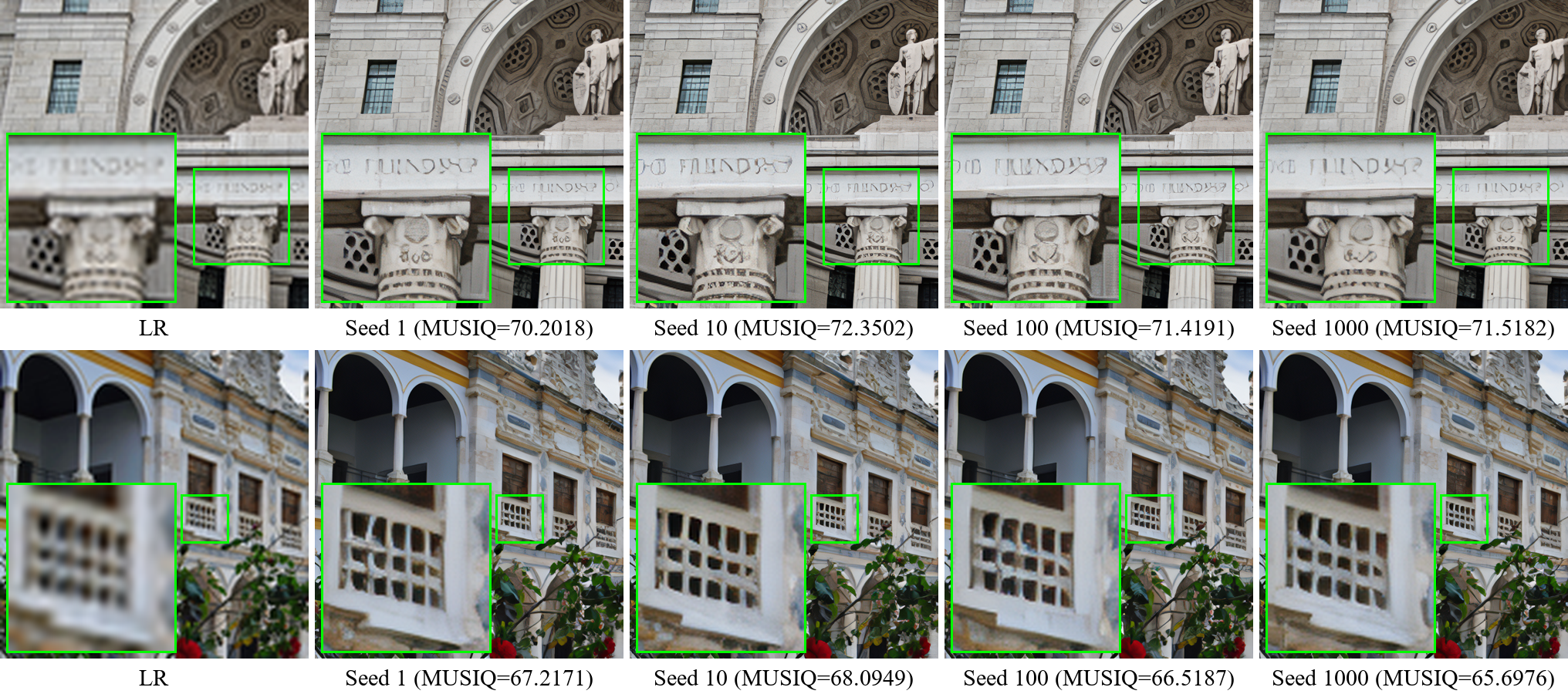} 
\caption{Qualitative comparison of SR results generated with different random seeds. (Zoom in for best view)}
\label{fig:inference_steps_comparison}
\end{figure}

We first briefly review the residual-shifting diffusion framework (ResShift) \cite{NEURIPS2023_2ac2eac5}, which underlies our method. 
Let $\boldsymbol{x}_0$ and $\boldsymbol{y}_0$ denote the latent codes of the HR and LR images encoded by the pretrained VQGAN \cite{Esser_2021_CVPR}, and let $\boldsymbol{e}_0 = \boldsymbol{y}_0 - \boldsymbol{x}_0$ denote their residual. Let $\{\eta_t\}_{t=1}^T$ be a monotonically increasing shifting sequence with $\eta_1 \to 0$ and $\eta_T \to 1$.

\textbf{Forward process.} ResShift constructs a Markov chain of length $T$ that gradually shifts the residual $\boldsymbol{e}_0$, whose transition distribution is formulated as
\begin{equation}
q(\boldsymbol{x}_t \mid \boldsymbol{x}_{t-1}, \boldsymbol{y}_0) = \mathcal{N}\left(\boldsymbol{x}_t;\; \boldsymbol{x}_{t-1} + \alpha_t\, \boldsymbol{e}_0,\; \kappa^2 \alpha_t\, \boldsymbol{I}\right),
\end{equation}
where $\alpha_t = \eta_t - \eta_{t-1}$ for $t > 1$ and $\alpha_1 = \eta_1$, and the hyperparameter $\kappa$ controls the noise variance: each step shifts away an $\alpha_t$ portion of the residual while injecting $\kappa^2 \alpha_t$ of Gaussian noise. 
Notably, the marginal distribution at any time step is analytically integrable:
\begin{equation}
\label{eq:margin forward distribution}
q(\boldsymbol{x}_t \mid \boldsymbol{x}_0, \boldsymbol{y}_0) = \mathcal{N}\left(\boldsymbol{x}_t;\; \boldsymbol{x}_0 + \eta_t\, \boldsymbol{e}_0,\; \kappa^2 \eta_t\, \boldsymbol{I}\right),
\end{equation}
namely the noisy sample deterministically interpolates from $\boldsymbol{x}_0$ toward $\boldsymbol{y}_0$ with a monotonically increasing noise level.

\textbf{Reverse process.} The reverse process starts from $p(\boldsymbol{x}_T \mid \boldsymbol{y}_0) \approx \mathcal{N}\left(\boldsymbol{x}_T;\; \boldsymbol{y}_0,\; \kappa^2 \boldsymbol{I}\right)$ and iteratively recovers $\boldsymbol{x}_0$ from $\boldsymbol{x}_t$ under the guidance of $\boldsymbol{y}_0$, with the inverse transition kernel assumed to be Gaussian:
\begin{equation}
p_\theta(\boldsymbol{x}_{t-1} \mid \boldsymbol{x}_t, \boldsymbol{y}_0) = \mathcal{N}\left(\boldsymbol{x}_{t-1};\; \boldsymbol{\mu}_\theta(\boldsymbol{x}_t, \boldsymbol{y}_0, t),\; \boldsymbol{\Sigma}_\theta(\boldsymbol{x}_t, \boldsymbol{y}_0, t)\right).
\end{equation}

\textbf{Training objective.} The optimization for $\theta$ is achieved by minimizing the Kullback-Leibler (KL) divergence between the forward posterior and the model reverse distribution, i.e., the negative evidence lower bound:
\begin{equation}
\min_{\theta}\; \sum_{t} D_{KL}\big[\, q(\boldsymbol{x}_{t-1} \mid \boldsymbol{x}_t, \boldsymbol{x}_0, \boldsymbol{y}_0)\; \big\|\; p_\theta(\boldsymbol{x}_{t-1} \mid \boldsymbol{x}_t, \boldsymbol{y}_0)\, \big],
\end{equation}
where the targeted distribution is rendered tractable by combining the forward transition and the marginal in Eq.~\eqref{eq:margin forward distribution}:
\begin{equation}
\label{eq:forward posterior}
q(\boldsymbol{x}_{t-1} \mid \boldsymbol{x}_t, \boldsymbol{x}_0, \boldsymbol{y}_0) = \mathcal{N}\left(\boldsymbol{x}_{t-1};\; \frac{\eta_{t-1}}{\eta_t}\, \boldsymbol{x}_t + \frac{\alpha_t}{\eta_t}\, \boldsymbol{x}_0,\; \kappa^2 \frac{\eta_{t-1}}{\eta_t}\, \alpha_t\, \boldsymbol{I}\right).
\end{equation}
Considering that the variance parameter is independent of $\boldsymbol{x}_t$ and $\boldsymbol{y}_0$, it is set to $\boldsymbol{\Sigma}_\theta = \Sigma_\theta\, \boldsymbol{I} = \kappa^2 \frac{\eta_{t-1}}{\eta_t}\, \alpha_t\, \boldsymbol{I}$, and the mean parameter is reparameterized as
\begin{equation}
\boldsymbol{\mu}_\theta(\boldsymbol{x}_t, \boldsymbol{y}_0, t) = \frac{\eta_{t-1}}{\eta_t}\, \boldsymbol{x}_t + \frac{\alpha_t}{\eta_t}\, \boldsymbol{f}_\theta(\boldsymbol{x}_t, \boldsymbol{y}_0, t),
\end{equation}
where $\boldsymbol{f}_\theta$ is a deep neural network aiming to predict $\boldsymbol{x}_0$. The KL objective then simplifies to the regression form $\min_{\theta} \sum_{t} w_t \left\| \boldsymbol{f}_\theta(\boldsymbol{x}_t, \boldsymbol{y}_0, t) - \boldsymbol{x}_0 \right\|_2^2$, where $w_t = \frac{\alpha_t}{2 \kappa^2 \eta_t \eta_{t-1}}$ is the step-dependent weight.

\textbf{Reverse sampling.} At inference, the reverse process is instantiated as the stochastic iteration:
\begin{equation}
\label{eq:reverse iteration}
\boldsymbol{x}_{t-1} = \boldsymbol{\mu}_\theta(\boldsymbol{x}_t, \boldsymbol{y}_0, t) + \sqrt{\Sigma_\theta}\; \boldsymbol{z}_{t-1} = \frac{\eta_{t-1}}{\eta_t}\, \boldsymbol{x}_t + \frac{\alpha_t}{\eta_t}\, \boldsymbol{x}_0^{\prime} + \kappa\sqrt{\frac{\eta_{t-1}\,\alpha_t}{\eta_t}}\; \boldsymbol{z}_{t-1},
\end{equation}
where $\boldsymbol{x}_0^{\prime} = \boldsymbol{f}_\theta(\boldsymbol{x}_t, \boldsymbol{y}_0, t)$ is the predicted clean image, and $\boldsymbol{z}_{t-1} \sim \mathcal{N}(\mathbf{0}, \boldsymbol{I})$ is the intermediate noise injected at step $t$. For $t = 1$, $\boldsymbol{z}_0 = \mathbf{0}$.

Our core observation is that, given the starting point $\boldsymbol{x}_T \sim \mathcal{N}(\boldsymbol{y}_0, \kappa^2 \boldsymbol{I})$, the quality of the generated image is dictated by the noise maps $\{\boldsymbol{z}_{t-1}\}_{t=2}^{T}$ sampled at the intermediate steps. To validate this observation, we fix the starting point $\boldsymbol{x}_T$ and perform inference with different random seeds, so that only the intermediate noise maps $\{\boldsymbol{z}_{t-1}\}_{t=2}^{T}$ differ across runs. The results are presented in Fig.~\ref{fig:inference_steps_comparison}, where the generated images are evaluated by the MUSIQ score \cite{ke2021musiq}. It can be observed that the image quality varies considerably across different random seeds, indicating that the noise maps sampled at the intermediate steps directly affect the quality of the generated images.

With the above observation, a natural question arises: does there exist a noise sampling distribution better than $\mathcal{N}(\mathbf{0}, \boldsymbol{I})$ that can improve the quality of the generated images? We will analyze this question in Sec.~\ref{sec:Method}.

\subsection{Method}
\label{sec:Method}
\begin{figure*}[t]
\centering
\includegraphics[width=\textwidth]{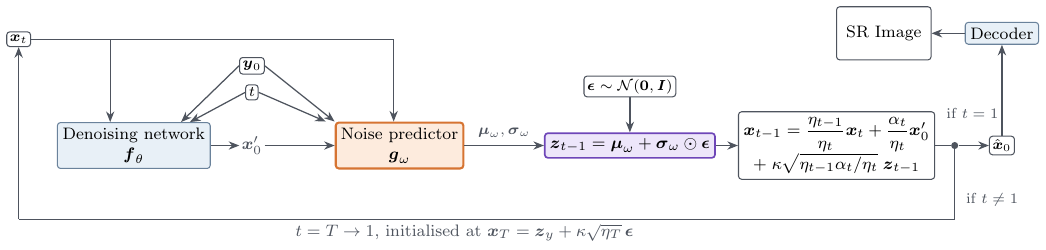}
\caption{Overview of the LPNSR inference framework. The noise predictor $\boldsymbol{g}_\omega$ outputs a diagonal Gaussian $(\boldsymbol{\mu}_\omega,\boldsymbol{\sigma}_\omega)$ whose sample $\boldsymbol{z}_{t-1}$ replaces the unconstrained Gaussian noise in the reverse iteration of Eq.~\eqref{eq:reverse iteration}. The reverse process runs from $t=T$ to $t=1$, and the final estimate is decoded by the VQGAN decoder into the SR image.}
\label{fig:overview}
\end{figure*}

As revealed in Sec.~\ref{sec:Motivation}, the noise maps sampled at the intermediate steps dictate the quality of the final generated image. 
Motivated by this observation, we parameterize the noise distribution at each intermediate step as a Gaussian distribution with diagonal covariance, whose mean and standard deviation are predicted element-wise by a learnable noise predictor $\boldsymbol{g}_\omega$:
\begin{equation}
\boldsymbol{z}_{t-1} = \boldsymbol{g}_\omega(\boldsymbol{x}_t, \boldsymbol{x}_0^{\prime}, \boldsymbol{y}_0, t) = \boldsymbol{\mu}_\omega(\boldsymbol{x}_t, \boldsymbol{x}_0^{\prime}, \boldsymbol{y}_0, t) + \boldsymbol{\sigma}_\omega(\boldsymbol{x}_t, \boldsymbol{x}_0^{\prime}, \boldsymbol{y}_0, t) \odot \boldsymbol{\varepsilon}_{t-1}, \qquad \boldsymbol{\varepsilon}_{t-1} \sim \mathcal{N}(\mathbf{0}, \boldsymbol{I}),
\end{equation}
where $\boldsymbol{x}_0^{\prime} = \boldsymbol{f}_\theta(\boldsymbol{x}_t, \boldsymbol{y}_0, t)$ is the clean-image prediction of the frozen denoiser. 
In the following, we present the two training schemes of $\boldsymbol{g}_\omega$, together with a per-step optimality analysis that underlies them. The overall inference framework is illustrated in Fig.~\ref{fig:overview}.

\textbf{End-to-end supervised training.} A natural idea following this observation is to train the noise predictor $\boldsymbol{g}_\omega$ with the quality of the final generated image as the supervision signal. 
To this end, $\boldsymbol{g}_\omega$ is injected into every intermediate step of the reverse chain, and the entire chain is unrolled end-to-end: starting from $\boldsymbol{x}_T = \boldsymbol{y}_0 + \kappa\,\boldsymbol{\epsilon}$, each reverse step computes $\boldsymbol{x}_0^{\prime} = \boldsymbol{f}_\theta(\boldsymbol{x}_t, \boldsymbol{y}_0, t)$ with the frozen denoiser, injects the noise predicted by $\boldsymbol{g}_\omega$, and proceeds via Eq.~\eqref{eq:reverse iteration}. 
Gradients from the final output are back-propagated through the whole unrolled chain to $\boldsymbol{g}_\omega$. 
Note that such training is computationally expensive, as it requires unrolling the entire reverse chain and back-propagating through every intermediate step, with both the memory and computation costs growing linearly with the number of sampling steps. 
The original ResShift framework \cite{NEURIPS2023_2ac2eac5} requires only four inference steps, which makes this training feasible; however, it may not be feasible for diffusion frameworks with considerably more inference steps. 
The training objective directly supervises the quality of the final generated image:
\begin{equation}
\mathcal{L} = \mathcal{L}_2\bigl(\hat{\boldsymbol{x}}_0,\, \boldsymbol{x}_0\bigr) \;+\; \lambda_l\, \mathcal{L}_{\mathrm{LPIPS}}\bigl(\mathcal{D}(\hat{\boldsymbol{x}}_0),\, \boldsymbol{x}_0\bigr) \;+\; \lambda_g\, \mathcal{L}_{\mathrm{GAN}}\bigl(\hat{\boldsymbol{x}}_0,\, \boldsymbol{x}_0\bigr),
\end{equation}
where $\hat{\boldsymbol{x}}_0$ denotes the final clean-image prediction produced at the last step of the unrolled chain, and $\mathcal{D}(\cdot)$ is the frozen VQGAN \cite{Esser_2021_CVPR} decoder that maps latents back to the image space. 
Here $\mathcal{L}_2$ is the regression loss in the latent space that constrains the fidelity of the generated latent, $\mathcal{L}_{\mathrm{LPIPS}}$ \cite{zhang2018unreasonable} is the perceptual loss that measures the perceptual similarity between the reconstructed and ground-truth images in the image space, and $\mathcal{L}_{\mathrm{GAN}}$ \cite{goodfellow2014generative} is the adversarial loss that distinguishes the generated latents from the real ones, encouraging realistic reconstructions. 
The complete training procedure is summarized in Alg.~\ref{alg:noise_predictor_training}.

\textbf{Mean-correction training.} 
We characterize the per-step optimal form of the injected noise from the perspective of distribution alignment: the injected noise is per-step optimal if it minimizes the KL divergence between the model reverse distribution and the forward posterior. 
By the noise parameterization above and the reparameterization in Eq.~\eqref{eq:reverse iteration}, the model reverse distribution becomes $\mathcal{N}\left(\boldsymbol{x}_{t-1};\, \boldsymbol{\mu}_\theta(\boldsymbol{x}_t, \boldsymbol{y}_0, t) + \sqrt{\Sigma_\theta}\, \boldsymbol{\mu}_\omega,\; \Sigma_\theta\,\mathrm{diag}(\boldsymbol{\sigma}_\omega^2)\right)$, while the forward posterior in Eq.~\eqref{eq:forward posterior} is Gaussian with mean $\tilde{\boldsymbol{\mu}}_t$ and covariance $\tilde{\boldsymbol{\Sigma}}_t$. 
The per-step optimal noise is thus obtained by minimizing this KL divergence:
\begin{equation}
\left(\boldsymbol{\mu}_\omega^{*},\; \boldsymbol{\sigma}_\omega^{*}\right) = \arg\min_{\boldsymbol{\mu}_\omega,\, \boldsymbol{\sigma}_\omega} D_{KL}\big(q(\boldsymbol{x}_{t-1} \mid \boldsymbol{x}_t, \boldsymbol{x}_0, \boldsymbol{y}_0)\;\big\|\; p_\theta(\boldsymbol{x}_{t-1} \mid \boldsymbol{x}_t, \boldsymbol{y}_0)\big).
\end{equation}
Since both distributions are Gaussian with diagonal covariance, minimizing this KL divergence simply amounts to aligning the mean and the variance of the two distributions.
Aligning the means yields the optimal mean prediction (the mean term is a non-negative quadratic in $\boldsymbol{\mu}_\omega$, minimized exactly when the two means coincide):
\begin{equation}
\label{eq:optimal noise for resshift}
\boldsymbol{\mu}_\omega^{*} = \frac{\tilde{\boldsymbol{\mu}}_t - \boldsymbol{\mu}_\theta(\boldsymbol{x}_t, \boldsymbol{y}_0, t)}{\sqrt{\Sigma_\theta}} = \frac{\alpha_t\,\bigl(\boldsymbol{x}_0 - \boldsymbol{x}_0^{\prime}\bigr)}{\eta_t\,\sqrt{\Sigma_\theta}},
\end{equation}
namely the optimal mean shift is precisely the normalized prediction residual of the denoising network. 
For the variance, since the reverse variance of ResShift \cite{NEURIPS2023_2ac2eac5} inherits the forward posterior variance, i.e., $\tilde{\boldsymbol{\Sigma}}_t = \boldsymbol{\Sigma}_\theta$, setting $\boldsymbol{\sigma}_\omega^{*} = \mathbf{1}$ suffices for the variance alignment. 
Substituting the per-step optimal mean $\boldsymbol{\mu}_\omega^{*}$ and variance scale $\boldsymbol{\sigma}_\omega^{*} = \mathbf{1}$ back into the model reverse distribution yields
\begin{equation}
p_\theta(\boldsymbol{x}_{t-1}\mid \boldsymbol{x}_t, \boldsymbol{y}_0) = \mathcal{N}\left(\boldsymbol{x}_{t-1};\, \boldsymbol{\mu}_\theta(\boldsymbol{x}_t, \boldsymbol{y}_0, t) + \frac{\alpha_t}{\eta_t}(\boldsymbol{x}_0 - \boldsymbol{x}_0'),\, \Sigma_\theta\,\boldsymbol{I}\right).
\end{equation}
This indicates that the per-step optimal injection is equivalent to rewriting the mean of $p_\theta$—that is, adding a mean-correction term on top of the frozen prediction. 
Formally, we express the reverse process as
\begin{equation}
p_{\theta,\omega}(\boldsymbol{x}_{t-1} \mid \boldsymbol{x}_t, \boldsymbol{y}_0) = \mathcal{N}\left(\boldsymbol{x}_{t-1};\; \boldsymbol{\mu}_{\theta,\omega}(\boldsymbol{x}_t, \boldsymbol{y}_0, t),\; \boldsymbol{\Sigma}_\theta\right),
\end{equation}
where the mean parameter is reparameterized as
\begin{equation}
\boldsymbol{\mu}_{\theta,\omega}(\boldsymbol{x}_t, \boldsymbol{y}_0, t) = \boldsymbol{\mu}_\theta(\boldsymbol{x}_t, \boldsymbol{y}_0, t) + \frac{\alpha_t}{\eta_t}\, \boldsymbol{f}_\omega(\boldsymbol{x}_t, \boldsymbol{x}_0^{\prime}, \boldsymbol{y}_0, t).
\end{equation}
Here $\boldsymbol{f}_\omega$ denotes the residual predictor, and the mean term of the injected noise is given by $\boldsymbol{\mu}_\omega = \frac{\alpha_t}{\eta_t\sqrt{\Sigma_\theta}}\,\boldsymbol{f}_\omega$, i.e., the frozen mean $\boldsymbol{\mu}_\theta$ is corrected by the residual prediction of $\boldsymbol{f}_\omega$. 
The optimization for $\omega$ is achieved by minimizing the negative evidence lower bound, namely:
\begin{equation}
\label{eq:mean correction objective}
\min_{\omega}\; \sum_{t} D_{KL}\big(q(\boldsymbol{x}_{t-1} \mid \boldsymbol{x}_t, \boldsymbol{x}_0, \boldsymbol{y}_0)\;\big\|\; p_{\theta,\omega}(\boldsymbol{x}_{t-1} \mid \boldsymbol{x}_t, \boldsymbol{y}_0)\big).
\end{equation}
The detailed mathematical derivation of this objective follows that of DDPM \cite{NEURIPS2020_4c5bcfec}.
Following ResShift \cite{NEURIPS2023_2ac2eac5}, we transform the optimization objective in Eq.~\eqref{eq:mean correction objective} into the following form:
\begin{equation}
\label{eq:mean correction loss}
\min_{\omega}\; \sum_{t} \bigl\|\, \boldsymbol{f}_\omega(\cdot) - (\boldsymbol{x}_0 - \boldsymbol{x}_0^{\prime}) \,\bigr\|_2^2 + \mathcal{L}_{\mathrm{LPIPS}}\bigl(\mathcal{D}(\boldsymbol{x}_0^{\prime} + \boldsymbol{f}_\omega(\cdot)),\, \boldsymbol{x}_0\bigr),
\end{equation}
where we incorporate the LPIPS perceptual loss \cite{zhang2018unreasonable} as a regularization term. As suggested in ResShift \cite{NEURIPS2023_2ac2eac5}, this term improves the perceptual quality of the generated images.

\begin{figure}
\centering
\begin{subfigure}{\linewidth}
\centering
\includegraphics[width=\linewidth, keepaspectratio]{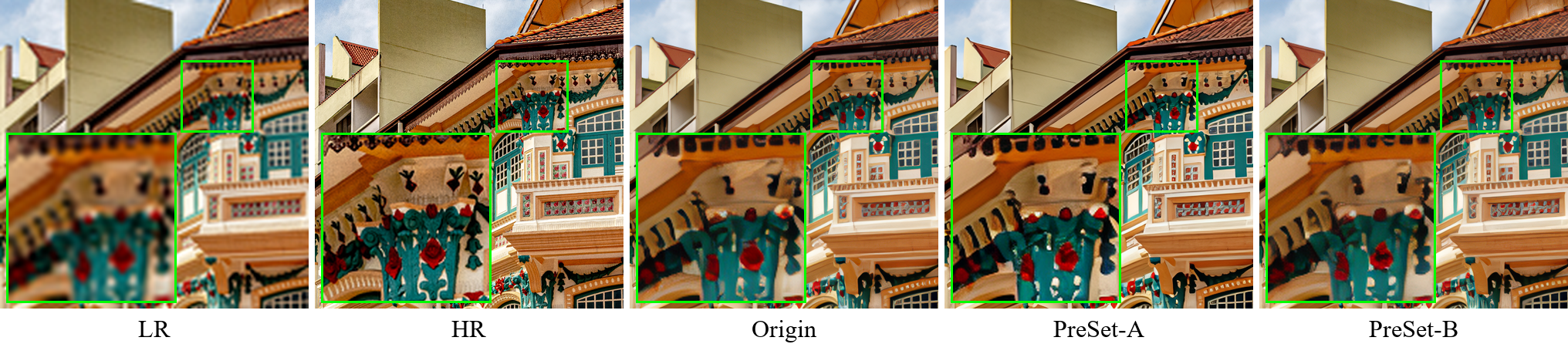} 
\end{subfigure}
\vspace{0.5em} 
\begin{subfigure}{\linewidth}
\centering
\includegraphics[width=\linewidth, keepaspectratio]{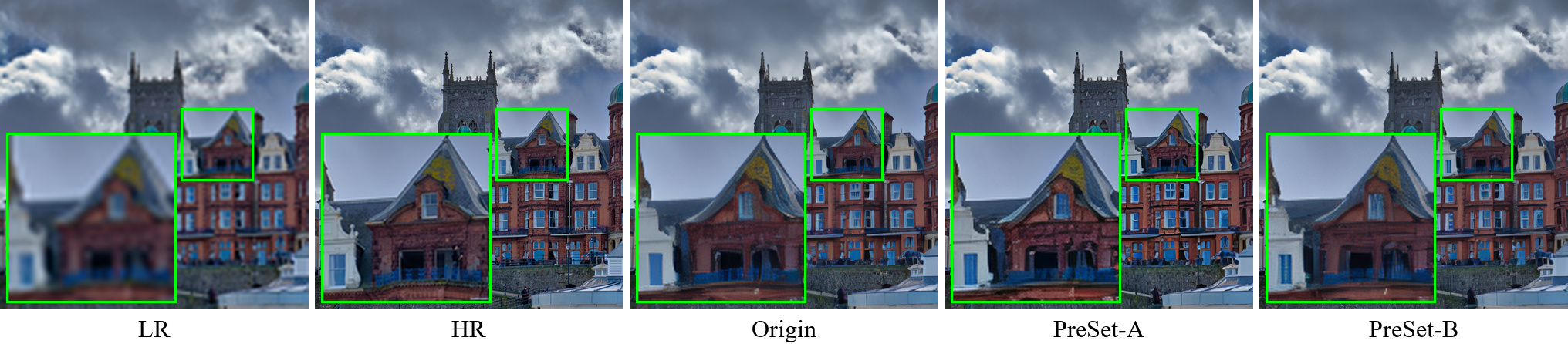} 
\end{subfigure}
\caption{Qualitative comparison of the SR results generated under different noise sampling schemes. (Zoom in for best view)}
\label{fig:model_comparison}
\end{figure}

\begin{table}[t]
\centering
\caption{Quantitative comparison on the LSDIR and ImageNet datasets. The last three rows of each group report our noise sampling schemes (Origin, LPNSR-A, and LPNSR-B), and the remaining rows list several representative SR methods for reference. The best and second-best results of each metric are highlighted in \textcolor{red}{red} and \textcolor{blue}{blue}, respectively.}
\label{tab:seed_comparison}
\resizebox{\linewidth}{!}{
\begin{tabular}{ll cccccccc}
\toprule
Dataset & Method & PSNR$\uparrow$ & SSIM$\uparrow$ & LPIPS$\downarrow$ & DISTS$\downarrow$ & MUSIQ$\uparrow$ & MANIQA$\uparrow$ & CLIPIQA$\uparrow$ & FID$\downarrow$ \\
\midrule
\multirow{7}{*}{\textit{LSDIR}} & VOSR & 20.04 & 0.4917 & 0.2448 & 0.1524 & \textcolor{blue}{72.1925} & 0.5025 & 0.6303 & 47.79 \\
 & InvSR & 19.19 & 0.4845 & 0.2657 & 0.1636 & \textcolor{red}{72.7310} & \textcolor{blue}{0.5113} & \textcolor{blue}{0.7156} & 69.30 \\
 & OSEDiff & 20.03 & 0.5094 & 0.2675 & 0.1627 & 71.4740 & 0.4573 & 0.6946 & 58.98 \\
 & SinSR & 20.39 & 0.5136 & 0.2281 & 0.1447 & 69.1867 & 0.4648 & 0.6687 & 40.82 \\
 & ResShift (Origin) & \textcolor{blue}{20.70} & \textcolor{blue}{0.5281} & 0.2249 & 0.1420 & 67.4741 & 0.4363 & 0.6445 & \textcolor{blue}{36.76} \\
 & \textbf{LPNSR-A (Ours)} & 19.77 & 0.5140 & \textcolor{red}{0.2099} & \textcolor{red}{0.1336} & 71.7904 & \textcolor{red}{0.5618} & \textcolor{red}{0.7389} & 42.39 \\
 & \textbf{LPNSR-B (Ours)} & \textcolor{red}{20.80} & \textcolor{red}{0.5345} & \textcolor{blue}{0.2197} & \textcolor{blue}{0.1394} & 68.3811 & 0.4582 & 0.6722 & \textcolor{red}{36.12} \\
\midrule
\multirow{7}{*}{\textit{ImageNet}} & VOSR & 24.98 & 0.6721 & 0.2445 & 0.1556 & \textcolor{red}{72.3900} & \textcolor{red}{0.4913} & 0.6327 & 45.64 \\
 & InvSR & 23.98 & 0.6689 & 0.2527 & 0.1616 & \textcolor{blue}{71.8737} & 0.4702 & \textcolor{red}{0.7152} & 57.64 \\
 & OSEDiff & 24.67 & 0.6917 & 0.2462 & 0.1544 & 71.6051 & 0.4595 & 0.6789 & 50.32 \\
 & SinSR & 26.53 & 0.7170 & 0.2251 & 0.1520 & 67.2132 & 0.4371 & 0.6655 & 41.24 \\
 & ResShift (Origin) & \textcolor{blue}{27.25} & \textcolor{blue}{0.7397} & 0.2058 & 0.1501 & 65.5388 & 0.3928 & 0.6238 & \textcolor{blue}{34.91} \\
 & \textbf{LPNSR-A (Ours)} & 26.31 & 0.7337 & \textcolor{red}{0.2026} & \textcolor{red}{0.1384} & 69.9097 & \textcolor{blue}{0.4897} & \textcolor{blue}{0.7101} & 36.93 \\
 & \textbf{LPNSR-B (Ours)} & \textcolor{red}{27.45} & \textcolor{red}{0.7430} & \textcolor{blue}{0.2047} & \textcolor{blue}{0.1481} & 66.6361 & 0.4095 & 0.6476 & \textcolor{red}{34.31} \\
\bottomrule
\end{tabular}
}
\end{table}

\begin{table}[t]
\centering
\caption{Statistics of the three noise sampling schemes on the LSDIR dataset. We randomly sample 100 random seeds; for each seed, every metric is averaged over the test images to obtain a per-seed mean, and the mean and the standard deviation of these per-seed means across the 100 seeds are reported as mean $\pm$ std.}
\label{tab:seed_stats}
\resizebox{\linewidth}{!}{
\begin{tabular}{lccccc}
\toprule
Method & PSNR$\uparrow$ & LPIPS$\downarrow$ & DISTS$\downarrow$ & MUSIQ$\uparrow$ & CLIPIQA$\uparrow$ \\
\midrule
Origin & $20.6951 \pm 0.0066$ & $0.2239 \pm 0.0009$ & $0.1419 \pm 0.0003$ & $67.5117 \pm 0.0617$ & $0.6436 \pm 0.0024$ \\
\textbf{PreSet-A} & $19.7679 \pm 0.0017$ & $0.2103 \pm 0.0002$ & $0.1339 \pm 0.0002$ & $71.8248 \pm 0.0269$ & $0.7393 \pm 0.0008$ \\
\textbf{PreSet-B} & $20.7975 \pm 0.0072$ & $0.2195 \pm 0.0007$ & $0.1394 \pm 0.0003$ & $68.5293 \pm 0.0640$ & $0.6726 \pm 0.0019$ \\
\bottomrule
\end{tabular}
}
\end{table}

\begin{table}[t]
\centering
\caption{Decomposition of the KL divergence between the forward posterior and the model reverse distribution at each intermediate step ($t=4,3,2$); all values are computed per image and then averaged over the test images of the LSDIR dataset. Per Eq.~\eqref{eq:kl decomposition}, $D_{KL}$ is split into a mean term $D_{KL}^{\text{mean}}$ and a variance term $D_{KL}^{\text{var}}$, where a smaller value indicates a better alignment, and $\bar{\sigma}_\omega^2$ denotes the mean injected noise variance, i.e., the mean of the diagonal entries of $\boldsymbol{\Sigma}_\omega$.}
\label{tab:kl_decomposition}
\resizebox{\linewidth}{!}{
\begin{tabular}{l ccc ccc ccc}
\toprule
\multirow{2}{*}{Method} & \multicolumn{3}{c}{$t=4$} & \multicolumn{3}{c}{$t=3$} & \multicolumn{3}{c}{$t=2$} \\
\cmidrule(lr){2-4}\cmidrule(lr){5-7}\cmidrule(lr){8-10}
 & $D_{KL}^{\text{mean}}$ & $D_{KL}^{\text{var}}$ & $\bar{\sigma}_\omega^2$ & $D_{KL}^{\text{mean}}$ & $D_{KL}^{\text{var}}$ & $\bar{\sigma}_\omega^2$ & $D_{KL}^{\text{mean}}$ & $D_{KL}^{\text{var}}$ & $\bar{\sigma}_\omega^2$ \\
\midrule
Origin & 0.0611 & 0.0000 & 1.0000 & 0.1782 & 0.0000 & 1.0000 & 9.2725 & 0.0000 & 1.0000 \\
\textbf{PreSet-A} & 0.8988 & 5.8690 & $1.47\times10^{-4}$ & 0.4707 & 6.1599 & $6.97\times10^{-5}$ & 15.1685 & 5.4199 & $2.49\times10^{-4}$ \\
\textbf{PreSet-B} & 0.0577 & 0.0000 & 1.0000 & 0.1710 & 0.0000 & 1.0000 & 9.2190 & 0.0000 & 1.0000 \\
\bottomrule
\end{tabular}
}
\end{table}

\begin{table}[t]
  \renewcommand{\arraystretch}{1.15}
  \centering
  \caption{Quantitative ablation studies on the loss functions used for training PreSet-A and PreSet-B, wherein the hyperparameters $\lambda_l$ and $\lambda_g$ control the weight importance of the LPIPS loss and the GAN loss, respectively. The results are evaluated on the LSDIR dataset under the 4-step sampling setting, with the last two columns measured at the intermediate step $t=2$.}
  \label{tab:ablation_study_loss_function}
  \resizebox{\linewidth}{!}{
  \begin{tabular}{cccccccccc}
    \toprule
    \multirow{2}{*}{Method} & \multicolumn{2}{c}{Hyperparameters} & \multicolumn{7}{c}{Metrics} \\
    \cmidrule{2-10}
    & $\lambda_l$ (LPIPS loss) & $\lambda_g$ (GAN loss) & PSNR$\uparrow$ & LPIPS$\downarrow$ & DISTS$\downarrow$ & MUSIQ$\uparrow$ & CLIPIQA$\uparrow$ & $D_{KL}^{\text{mean}}$$\downarrow$ & $\bar{\sigma}_\omega^2$ \\
    \midrule
    A-Baseline1 & 0.0 & 0.0 & 22.05 & 0.2683 & 0.1592 & 64.2017 & 0.6184 & 8.8653 & $1.93\times10^{-4}$ \\
    A-Baseline2 & 1.0 & 0.0 & 20.35 & 0.2155 & 0.1372 & 70.4506 & 0.7180 & 12.3537 & $2.56\times10^{-4}$ \\
    A-Baseline3 & 0.0 & 0.1 & 19.28 & 0.2230 & 0.1428 & 73.0554 & 0.7495 & 17.0581 & $2.91\times10^{-4}$ \\
    B-Baseline1 & 0.0 & \textbackslash & 21.55 & 0.2602 & 0.1571 & 64.3558 & 0.6202 & 8.9522 & 1.0000 \\
    \textbf{PreSet-A} & 1.0 & 0.1 & 19.77 & 0.2099 & 0.1336 & 71.7904 & 0.7389 & 15.1685 & $2.49\times10^{-4}$ \\
    \textbf{PreSet-B} & 1.0 & \textbackslash & 20.80 & 0.2197 & 0.1394 & 68.3811 & 0.6722 & 9.2190 & 1.0000 \\
    \bottomrule
  \end{tabular}
  }
\end{table}

\section{Experiments}
\label{sec:Experiments}
In this section, we conduct experiments to validate the role of the noise predictor and compare the performance of different schemes. For clarity, we denote the standard Gaussian distribution serving as the noise sampling distribution as Origin (ResShift), the noise predictor trained with the end-to-end supervised training scheme as PreSet-A (LPNSR-A), and the one trained with the mean-correction training scheme as PreSet-B (LPNSR-B).
Following common practice in the field, our experiments mainly focus on the $\times 4$ SR task.

\subsection{Experimental setup}
\textbf{Training Details.} We collect images from public sources including LSDIR \cite{li2023lsdir}, DIV2K \cite{agustsson2017ntire}, and Flickr2K, as well as web images subjected to quality filtering, totaling 1M images. We train the noise predictor on the collected images for over 200k iterations, randomly cropping an image patch with a resolution of $256 \times 256$ from the source image and synthesizing the LR image using the pipeline of Real-ESRGAN \cite{wang2021real} at each iteration.
We adopt the AdamW \cite{loshchilov2017decoupled} optimizer with a learning rate of $5 \times 10^{-5}$ and a batch size of 16.
The hyperparameters for the loss function are set as $\lambda_l = 1.0$ and $\lambda_g = 0.1$.
During training, we set $T=4$ to remain consistent with ResShift \cite{NEURIPS2023_2ac2eac5}, and the noise variance hyperparameter $\kappa=2.0$ as well as the shifting sequence $\{\eta_t\}_{t=1}^T$ also follow the identical settings.
The denoising network $\boldsymbol{f}_\theta$ and the VQGAN autoencoder \cite{Esser_2021_CVPR} are frozen during training, and only the noise predictor is optimized.
The training is conducted on one NVIDIA RTX 5090 GPU. More detailed hyperparameter configurations are provided in the \textbf{Appendix}.

\textbf{Testing Datasets and Metrics.} For the testing datasets, we randomly sample 1000 images from the ImageNet test split \cite{deng2009imagenet}, whose HR images are randomly cropped to $512 \times 512$ resolution and then degraded through the Real-ESRGAN pipeline \cite{wang2021real} to produce $128 \times 128$ LR images.
For the LSDIR \cite{li2023lsdir} test split, we randomly crop the paired LR and HR images to $128 \times 128$ and $512 \times 512$, respectively.
We evaluate our method together with the competing methods with full-reference and no-reference metrics. For distortion fidelity, we report PSNR and SSIM \cite{wangzhou2004image} on the Y channel of the YCbCr color space. For reference-based perceptual quality, we use LPIPS \cite{zhang2018unreasonable}, DISTS \cite{ding2022image}, and FID \cite{heusel2017gans}. For no-reference perceptual quality, we report MUSIQ \cite{ke2021musiq}, MANIQA \cite{yang2022maniqa}, and CLIPIQA \cite{wang2023exploring}.

\textbf{Compared Methods.} We compare our method with five representative diffusion-based SR approaches: ResShift \cite{NEURIPS2023_2ac2eac5}, SinSR \cite{wang2024sinsr}, OSEDiff \cite{wu2024one}, InvSR \cite{Yue_2025_CVPR}, and VOSR \cite{wu2026vosr}. Among them, OSEDiff and InvSR are built upon large-scale pretrained text-to-image diffusion models. All methods are run with their official default settings.

\textbf{Network Architecture.} For the noise predictor $\boldsymbol{g}_\omega$, we follow the default architecture and parameter settings of ResShift \cite{NEURIPS2023_2ac2eac5}, i.e., the same UNet backbone in which the self-attention layers are replaced by Swin Transformer \cite{liu2021swin} blocks, so as to improve the robustness of the model to arbitrary image resolutions.
For the adversarial loss, we adopt the classical PatchGAN discriminator \cite{isola2017image} trained with the hinge loss \cite{lim2017geometric}.

\subsection{Experimental results}
\label{sec:Experimental Results}

\begin{figure}[t]
\centering
\begin{subfigure}{\linewidth}
\centering
\includegraphics[width=\linewidth, keepaspectratio]{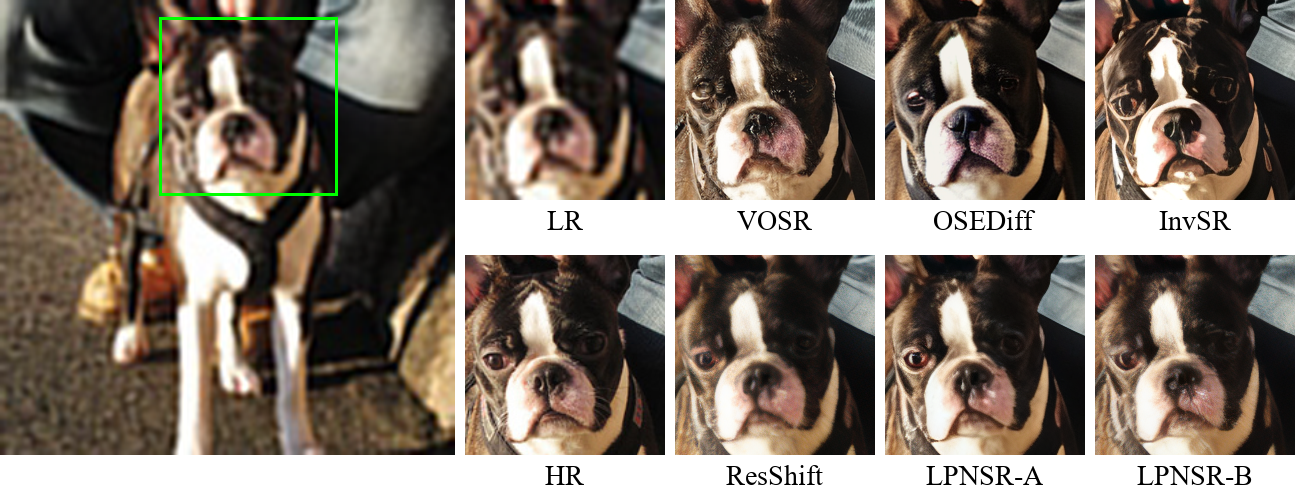} 
\end{subfigure}
\vspace{0.5em} 
\begin{subfigure}{\linewidth}
\centering
\includegraphics[width=\linewidth, keepaspectratio]{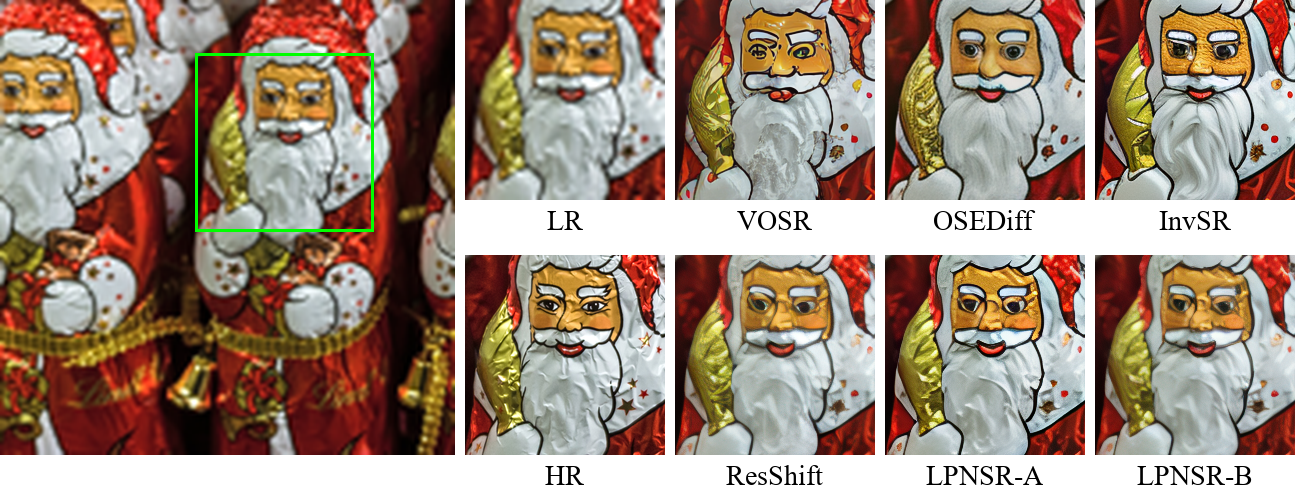} 
\end{subfigure}
\caption{Qualitative comparison with other methods. (Zoom in for best view)}
\label{fig:lpnsr_comparison}
\end{figure}

\textbf{Comparison with Other Methods.} Tab.~\ref{tab:seed_comparison} compares our noise sampling schemes with the compared methods on the LSDIR and ImageNet datasets. Our LPNSR-B ranks first in PSNR, SSIM, and FID on both datasets, while LPNSR-A ranks first in LPIPS and DISTS on both datasets and additionally in MANIQA and CLIPIQA on LSDIR. The compared methods lead only on MUSIQ (InvSR on LSDIR and VOSR on ImageNet) and, on ImageNet, on MANIQA and CLIPIQA (VOSR and InvSR, respectively). Notably, our schemes achieve these results without any text-to-image pretraining. The qualitative comparison in Fig.~\ref{fig:lpnsr_comparison} further corroborates these results. LPNSR-A and LPNSR-B restore sharper textures and more faithful details, such as the fine fur strands of the dog and the wrinkled foil surface of the Santa Claus. In contrast, VOSR, OSEDiff, and InvSR tend to hallucinate textures and details that are unfaithful to the original image (e.g., the facial features of the dog and of the Santa Claus figure), while ResShift over-smooths the results and thus yields an unclear visual appearance.

\textbf{Noise Sampling Scheme Comparison.} Based on the results in Tab.~\ref{tab:seed_comparison}, we compare the three noise sampling schemes on the LSDIR and ImageNet datasets, with Origin (ResShift) as the baseline. PreSet-A (LPNSR-A) substantially improves all perceptual metrics (LPIPS, DISTS, MUSIQ, MANIQA, and CLIPIQA) at the cost of degraded fidelity (PSNR, SSIM) and an enlarged FID, whereas PreSet-B (LPNSR-B), derived from the per-step mean-correction analysis in Sec.~\ref{sec:Method}, improves all metrics simultaneously---fidelity, perceptual quality, and FID---though its perceptual gains are less pronounced than those of PreSet-A. Fig.~\ref{fig:model_comparison} corroborates these behaviors qualitatively: Origin tends to produce over-smoothed images lacking high-frequency details, PreSet-A recovers richer textures at the cost of slight structural deviation, and PreSet-B achieves a visually appealing balance. Tab.~\ref{tab:seed_stats} further reports the mean and standard deviation over 100 random seeds on the LSDIR dataset. The standard deviations of PreSet-A are consistently and substantially smaller than those of the other two schemes, whereas Origin and PreSet-B fluctuate at similar levels---as expected, since PreSet-B only corrects the mean term of the reverse process, whereas the noise predictor of PreSet-A additionally predicts and corrects the variance term. Overall, the deviations are remarkably small, confirming that the above observations are not artifacts of a particular seed.

\textbf{Distribution Alignment Analysis.} We further measure the KL divergence between the forward posterior $q(\boldsymbol{x}_{t-1}\mid \boldsymbol{x}_t, \boldsymbol{x}_0, \boldsymbol{y}_0)$ and the model reverse distribution $p_\theta(\boldsymbol{x}_{t-1}\mid \boldsymbol{x}_t, \boldsymbol{y}_0)$ at each intermediate step ($t=4,3,2$). Both are multivariate Gaussians built from the ground-truth $\boldsymbol{x}_0$ (posterior) and the denoised estimate $\boldsymbol{x}_0'$ (reverse distribution), so their divergence follows the standard closed-form solution and decouples into a mean term and a variance term:
\begin{equation}
\label{eq:kl decomposition}
D_{KL} = \underbrace{\frac{1}{2}\,\boldsymbol{\Delta}^{\!\top}\Sigma_\theta^{-1}\boldsymbol{\Delta}}_{D_{KL}^{\text{mean}}} + \underbrace{\frac{1}{2}\left[\mathrm{tr}\left(\boldsymbol{\Sigma}_\omega - \boldsymbol{I}\right) - \ln\det\left(\boldsymbol{\Sigma}_\omega\right)\right]}_{D_{KL}^{\text{var}}},
\end{equation}
where $\boldsymbol{\Sigma}_\omega = \mathrm{diag}(\boldsymbol{\sigma}_\omega^2)$ denotes the diagonal covariance of the injected noise and $\boldsymbol{\Delta} = \tilde{\boldsymbol{\mu}}_t - \boldsymbol{\mu}_\theta - \sqrt{\Sigma_\theta}\,\boldsymbol{\mu}_\omega$ is the mean deviation between the two distributions.
The divergence is computed as $D_{KL}(p \,\|\, q)$ rather than $D_{KL}(q \,\|\, p)$, since the variance term of the latter is governed by $\mathrm{tr}(\boldsymbol{\Sigma}_\omega^{-1})$ and thus varies by orders of magnitude across the compared schemes, which makes the resulting values not directly comparable.
Tab.~\ref{tab:kl_decomposition} reports the decomposition at each intermediate step. PreSet-A, which achieves the best perceptual quality (Tab.~\ref{tab:seed_comparison}), deviates by far the most in its mean term from the forward posterior mean, which explains its degradation in PSNR and SSIM and reveals a general phenomenon that the ground-truth image does not necessarily represent the best perceptual quality. By contrast, PreSet-B stays closer to the forward posterior mean than Origin, which accounts for its gains in the fidelity metrics, and its improved perceptual quality can be attributed to approaching the perceptual quality ceiling of the ground-truth image. Moreover, the injected noise variance of PreSet-A is orders of magnitude smaller ($\bar{\sigma}_\omega^2 \sim 10^{-4}$ versus $1$), consistent with its markedly lower standard deviations across seeds (Tab.~\ref{tab:seed_stats}). The mean term generally increases as the step $t$ decreases, since $\boldsymbol{\Delta}$ is scaled by $\alpha_t/\eta_t$, which grows when $t$ decreases. 

\textbf{Loss Function Ablation.} We ablate the loss functions for training the noise predictors of PreSet-A and PreSet-B, as reported in Tab.~\ref{tab:ablation_study_loss_function}. 
For PreSet-B, removing the LPIPS loss (B-Baseline1) improves the fidelity but severely degrades the perceptual quality, indicating that the LPIPS loss is essential for its perceptual quality. 
For PreSet-A, a similar trend is observed: the L2 loss alone (A-Baseline1) yields the highest pixel fidelity but the weakest perceptual quality, the LPIPS loss (A-Baseline2) markedly improves the perceptual quality at a small cost of fidelity, and the GAN loss (A-Baseline3) further enhances the no-reference realism while degrading the full-reference fidelity. Combining both losses (PreSet-A) achieves the best overall trade-off between pixel-level fidelity and perceptual quality.

\textbf{Additional Results.} More experimental results are provided in the \textbf{Appendix}.

\section{Conclusion}
\label{sec:Conclusion}
In this paper, we started from the observation that, once the starting point of the reverse chain and the denoising network are fixed, the quality of the generated image is dictated by the noise maps sampled at the intermediate steps.
We therefore parameterized the intermediate noise sampling distribution as a learnable conditional distribution, instantiated it on the residual-shifting diffusion paradigm, and proposed LPNSR, in which a multi-input-aware noise predictor estimates the mean and the variance of the injected noise at each sampling step while leaving the pretrained denoising network and the residual-shifting mechanism untouched.
We further developed two training schemes, i.e., the end-to-end supervised training and the mean-correction training, and analyzed the per-step distribution gap to explain their complementary behaviors in fidelity and perceptual quality.
Extensive experiments validate the effectiveness of the learnable noise sampling distribution, showing that LPNSR improves the generated image quality over the standard Gaussian distribution.

\textbf{Future Work.} Applying the learnable noise sampling distribution to diffusion frameworks with richer generative priors, e.g., large-scale pretrained text-to-image models, is expected to further improve the quality of the generated images. Moreover, designing alternative training schemes for the noise predictor is another promising direction for future work.

{
\small
\bibliographystyle{unsrt} 
\bibliography{neurips_2026_references} 
}


\newpage
\appendix

\section{Appendix}
In the appendix, we provide the following materials:
\begin{itemize}
    \item The hyperparameter configuration of the noise predictor, including the model architecture and the training hyperparameters.
    \item The complete training algorithms of our LPNSR framework.
    \item The inference efficiency analysis of the three noise sampling schemes.
    \item A comparison between the noise sampling distribution and a deterministic noise map under the end-to-end supervised training scheme.
    \item More qualitative comparisons of the SR results.
\end{itemize}

\subsection{Hyperparameter Configuration}
\label{sec:hyperparameter_config}
The architecture of the noise predictor $\boldsymbol{g}_\omega$ follows the default setting of ResShift \cite{NEURIPS2023_2ac2eac5}. The key model and training hyperparameters are summarized in Tabs.~\ref{tab:model_config} and~\ref{tab:training_config}.

\begin{table}[h]
\centering
\caption{Architecture hyperparameters of the noise predictor $\boldsymbol{g}_\omega$.}
\label{tab:model_config}
\begin{tabular}{ll}
\toprule
Hyperparameter & Value \\
\midrule
Base channels & 160 \\
Channel multipliers & $(1, 2, 2, 4)$ \\
Residual blocks per resolution & 2 \\
Attention resolutions & $64, 32, 16, 8$ \\
Channels per attention head & 32 \\
Scale-shift normalization & Enabled \\
Swin Transformer depth & 2 \\
Swin Transformer embedding dimension & 192 \\
Swin Transformer window size & 8 \\
MLP ratio & 4.0 \\
Dropout & 0.0 \\
LR latent resolution & $64 \times 64$ \\
Latent channels & 3 \\
LR conditioning & Enabled \\
\bottomrule
\end{tabular}
\end{table}

\begin{table}[h]
\centering
\caption{Training hyperparameters of the noise predictor.}
\label{tab:training_config}
\begin{tabular}{ll}
\toprule
Hyperparameter & Value \\
\midrule
Training iterations & 200k \\
Batch size & 16 \\
Optimizer & AdamW \\
Learning rate & $5 \times 10^{-5}$ \\
Weight decay & 0 \\
LR schedule & None \\
EMA rate & 0.999 \\
Gradient clipping & 1.0 \\
Mixed precision (AMP) & Enabled \\
\cmidrule{1-2}
Sampling steps $T$ & 4 \\
Noise variance $\kappa$ & 2.0 \\
Shift schedule & Exponential \\
\cmidrule{1-2}
Loss weights ($\mathcal{L}_2$ / LPIPS / GAN) & $1.0$ / $1.0$ / $0.1$ \\
Discriminator & PatchGAN with hinge loss \\
\bottomrule
\end{tabular}
\end{table}

\subsection{Training Algorithms}
\label{sec:training_algorithms}
The end-to-end supervised training and the mean-correction training of the noise predictor are summarized in Alg.~\ref{alg:noise_predictor_training} and Alg.~\ref{alg:mean_correction_training}, respectively.

\begin{figure}[htbp]
  \begin{minipage}[t]{0.9\textwidth}
    \begin{algorithm}[H]
      \caption{End-to-End Noise Predictor Training}
      \label{alg:noise_predictor_training}
      \begin{algorithmic}[1]
        \Require HR/LR training pairs, frozen VQGAN $\mathcal{E}/\mathcal{D}$, frozen denoiser $\boldsymbol{f}_\theta$, noise predictor $\boldsymbol{g}_\omega$, total iterations $N$
        \Ensure Trained noise predictor $\boldsymbol{g}_\omega$
        \For{$iteration = 1, \dots, N$}
          \State Sample HR/LR images from the training set; \quad $\boldsymbol{x}_0 = \mathcal{E}(\text{HR image})$, $\boldsymbol{y}_0 = \mathcal{E}(\text{LR image})$; $\boldsymbol{x}_T = \boldsymbol{y}_0 + \kappa\,\boldsymbol{\epsilon}$, $\boldsymbol{\epsilon} \sim \mathcal{N}(\mathbf{0}, \boldsymbol{I})$
          \For{$t = T, T-1, \dots, 1$}
            \State $\boldsymbol{x}_0^{\prime} = \boldsymbol{f}_\theta(\boldsymbol{x}_t, \boldsymbol{y}_0, t)$
            \If{$t > 1$}
              \State $\boldsymbol{x}_{t-1} = \frac{\eta_{t-1}}{\eta_t}\, \boldsymbol{x}_t + \frac{\alpha_t}{\eta_t}\, \boldsymbol{x}_0^{\prime} + \kappa\sqrt{\frac{\eta_{t-1}\,\alpha_t}{\eta_t}}\; \boldsymbol{g}_\omega(\boldsymbol{x}_t, \boldsymbol{x}_0^{\prime}, \boldsymbol{y}_0, t)$
            \EndIf
            \State $\hat{\boldsymbol{x}}_0 = \boldsymbol{x}_0^{\prime}$
          \EndFor
          \State $\mathcal{L} = \mathcal{L}_2(\hat{\boldsymbol{x}}_0,\, \boldsymbol{x}_0) + \lambda_l\, \mathcal{L}_{\mathrm{LPIPS}}(\mathcal{D}(\hat{\boldsymbol{x}}_0),\, \boldsymbol{x}_0) + \lambda_g\, \mathcal{L}_{\mathrm{GAN}}(\hat{\boldsymbol{x}}_0,\, \boldsymbol{x}_0)$
          \State Update $\boldsymbol{g}_\omega$
        \EndFor
        \State \Return $\boldsymbol{g}_\omega$
      \end{algorithmic}
    \end{algorithm}
  \end{minipage}
\end{figure}

\begin{figure}[htbp]
  \begin{minipage}[t]{0.9\textwidth}
    \begin{algorithm}[H]
      \caption{Mean-Correction Noise Predictor Training}
      \label{alg:mean_correction_training}
      \begin{algorithmic}[1]
        \Require HR/LR training pairs, frozen VQGAN encoder/decoder $\mathcal{E}/\mathcal{D}$, frozen denoiser $\boldsymbol{f}_\theta$, noise predictor $\boldsymbol{f}_\omega$, total iterations $N$
        \Ensure Trained noise predictor $\boldsymbol{f}_\omega$
        \For{$iteration = 1, \dots, N$}
          \State Sample HR/LR images from the training set; \quad $\boldsymbol{x}_0 = \mathcal{E}(\text{HR image})$, $\boldsymbol{y}_0 = \mathcal{E}(\text{LR image})$; sample $t \sim \mathrm{Uniform}\{1, \dots, T\}$
          \State Forward: $\boldsymbol{x}_t = \boldsymbol{x}_0 + \eta_t\,(\boldsymbol{y}_0 - \boldsymbol{x}_0) + \kappa\sqrt{\eta_t}\,\boldsymbol{\epsilon}$, $\boldsymbol{\epsilon} \sim \mathcal{N}(\mathbf{0}, \boldsymbol{I})$
          \State $\boldsymbol{x}_0^{\prime} = \boldsymbol{f}_\theta(\boldsymbol{x}_t, \boldsymbol{y}_0, t)$
          \State $\boldsymbol{r} = \boldsymbol{f}_\omega(\boldsymbol{x}_t, \boldsymbol{x}_0^{\prime}, \boldsymbol{y}_0, t)$
          \State $\mathcal{L} = \| \boldsymbol{r} - (\boldsymbol{x}_0 - \boldsymbol{x}_0^{\prime}) \|_2^2 + \mathcal{L}_{\mathrm{LPIPS}}(\mathcal{D}(\boldsymbol{x}_0^{\prime} + \boldsymbol{r}),\, \boldsymbol{x}_0)$
          \State Update $\boldsymbol{f}_\omega$
        \EndFor
        \State \Return $\boldsymbol{f}_\omega$
      \end{algorithmic}
    \end{algorithm}
  \end{minipage}
\vspace{2em}
\end{figure}

\subsection{Efficiency Analysis}
\label{sec:efficiency_analysis}
Tab.~\ref{tab:efficiency} reports the inference efficiency of the three schemes. Since the noise predictor only estimates the mean and the variance of the injected noise at each of the 4 sampling steps, the extra cost is moderate, raising the latency and the GPU memory by about $1.4\times$ and $1.2\times$, respectively. PreSet-A and PreSet-B share the same architecture and hence nearly identical efficiency. Overall, the inference cost remains within an acceptable range.

\begin{table}[t]
\centering
\caption{Inference efficiency of the three noise sampling schemes on the LSDIR dataset ($128 \times 128$ LR $\rightarrow$ $512 \times 512$ SR), measured on a single NVIDIA RTX 5090 GPU with a batch size of 1 and mixed precision enabled. The number of parameters and the FLOPs are reported per image, and the latency and the GPU memory usage are averaged over all test images.}
\label{tab:efficiency}
\begin{tabular}{lcccc}
\toprule
Method & \#Params (M)$\downarrow$ & FLOPs (G)$\downarrow$ & Latency (ms)$\downarrow$ & GPU Memory Usage (MB)$\downarrow$ \\
\midrule
Origin & 173.92 & 5410.3 & 321.8 & 1968 \\
\textbf{PreSet-A} & 292.52 & 7032.3 & 442.1 & 2432 \\
\textbf{PreSet-B} & 292.51 & 7031.8 & 438.5 & 2432 \\
\bottomrule
\end{tabular}
\end{table}

\subsection{Distribution Prediction versus Deterministic Noise Map}
\label{sec:distribution_vs_deterministic}
Under the end-to-end supervised training scheme, the noise predictor admits two different instantiations. In the first one, the predictor estimates the mean and the variance of the noise sampling distribution at each intermediate step, so that the injected noise is sampled from the resulting conditional Gaussian. In the second one, the predictor directly outputs a deterministic noise map at each intermediate step, i.e., the stochastic term is removed and the predicted map itself is injected. Both instantiations are trained with exactly the same configuration. We compare the two instantiations in Tab.~\ref{tab:deterministic_noise}. Although the variance predicted by the distribution-based instantiation is orders of magnitude smaller than that of the standard Gaussian ($\bar{\sigma}_\omega^2 \sim 10^{-4}$, Tab.~\ref{tab:kl_decomposition} in the main paper), it still performs better than its deterministic counterpart on all the reported metrics, indicating that discarding the stochastic term entirely degrades the quality of the generated images.

\begin{table}[t]
\centering
\caption{Ablation on the form of the noise predictor under the end-to-end supervised training scheme, i.e., predicting the mean and the variance of the noise sampling distribution versus directly predicting a deterministic noise map. The results are evaluated on the LSDIR dataset under the 4-step sampling setting.}
\label{tab:deterministic_noise}
\resizebox{\linewidth}{!}{
\begin{tabular}{lccccccc}
\toprule
Method & PSNR$\uparrow$ & SSIM$\uparrow$ & LPIPS$\downarrow$ & DISTS$\downarrow$ & MUSIQ$\uparrow$ & MANIQA$\uparrow$ & CLIPIQA$\uparrow$ \\
\midrule
Deterministic noise map & 19.73 & 0.5174 & 0.2315 & 0.1456 & 70.0219 & 0.4491 & 0.6452 \\
Noise sampling distribution & 19.77 & 0.5140 & 0.2099 & 0.1336 & 71.7904 & 0.5618 & 0.7389 \\
\bottomrule
\end{tabular}
}
\end{table}

\subsection{More Qualitative Comparisons}
\label{sec:more_qualitative}
\begin{figure}[t]
\centering
\begin{subfigure}{\linewidth}
\centering
\includegraphics[width=\linewidth, keepaspectratio]{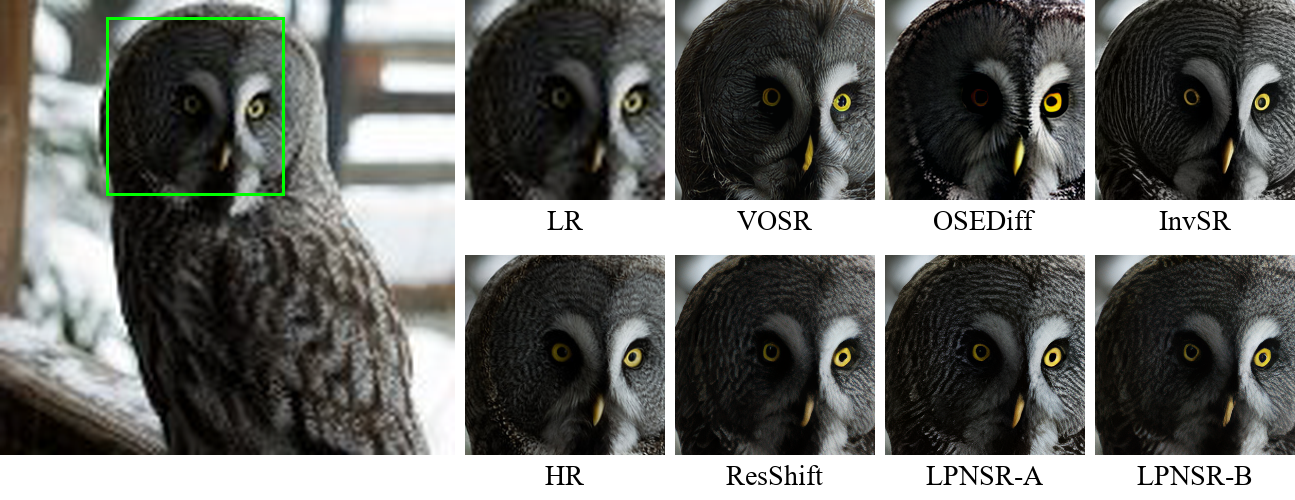}
\end{subfigure}
\vspace{0.5em}
\begin{subfigure}{\linewidth}
\centering
\includegraphics[width=\linewidth, keepaspectratio]{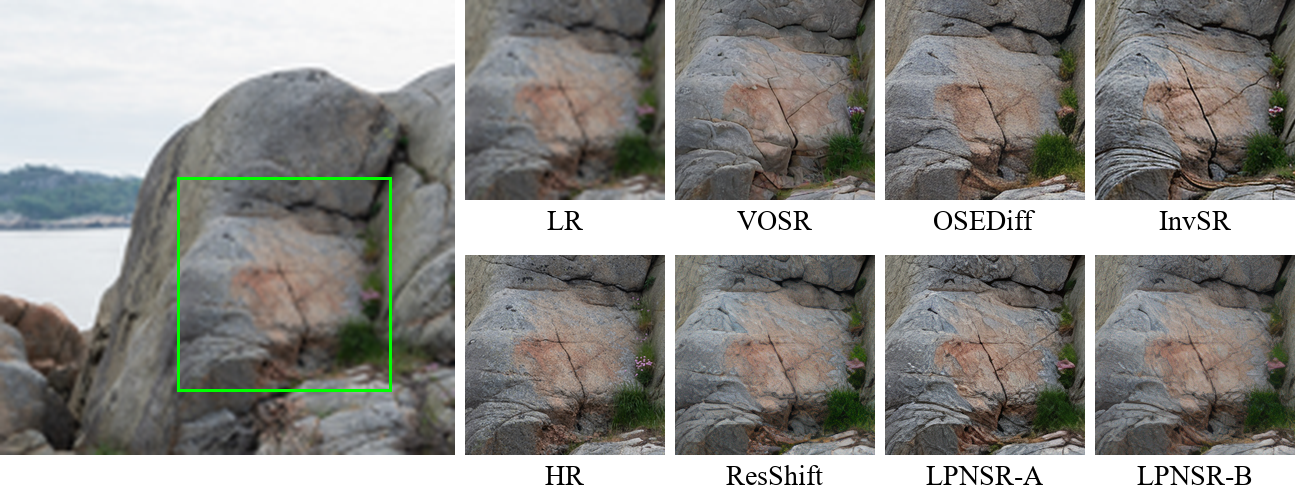}
\end{subfigure}
\vspace{0.5em}
\begin{subfigure}{\linewidth}
\centering
\includegraphics[width=\linewidth, keepaspectratio]{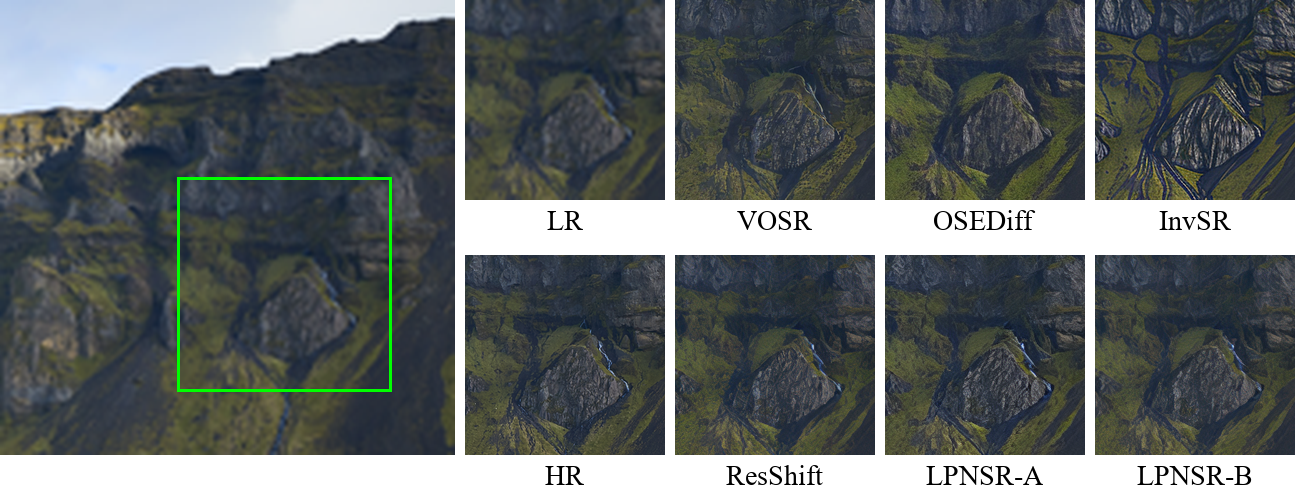}
\end{subfigure}
\caption{Additional qualitative comparison with other methods. (Zoom in for best view)}
\label{fig:model_comparison_3}
\end{figure}

\begin{figure}[t]
\centering
\begin{subfigure}{\linewidth}
\centering
\includegraphics[width=\linewidth, keepaspectratio]{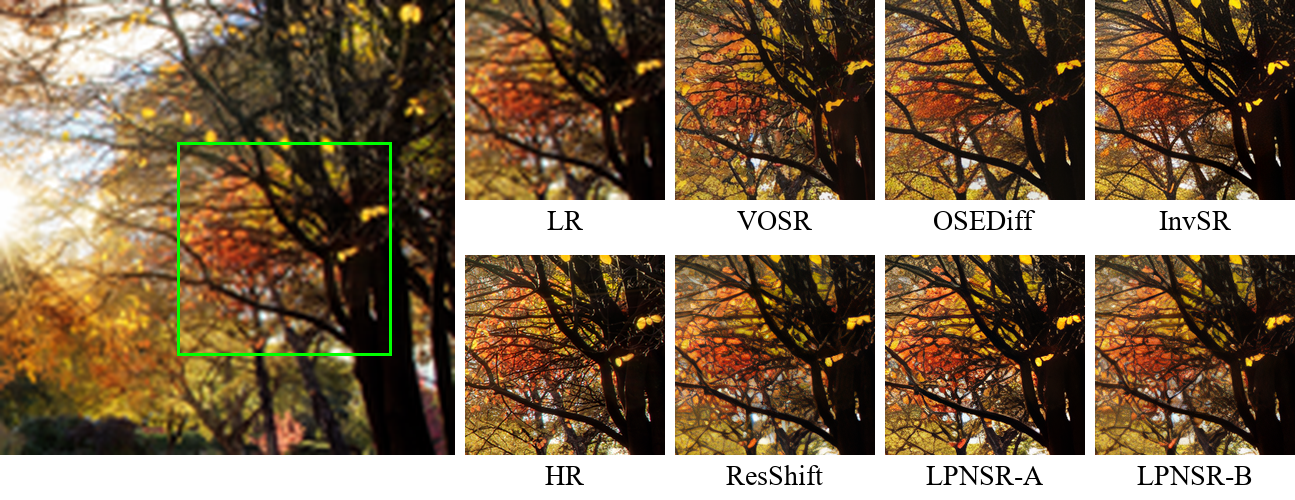}
\end{subfigure}
\vspace{0.5em}
\begin{subfigure}{\linewidth}
\centering
\includegraphics[width=\linewidth, keepaspectratio]{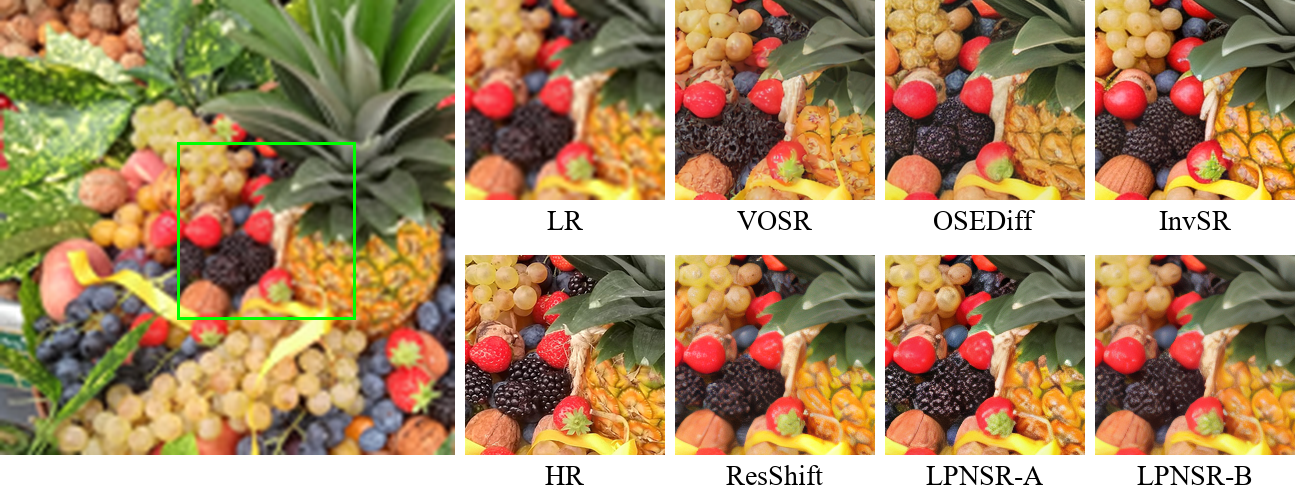}
\end{subfigure}
\vspace{0.5em}
\begin{subfigure}{\linewidth}
\centering
\includegraphics[width=\linewidth, keepaspectratio]{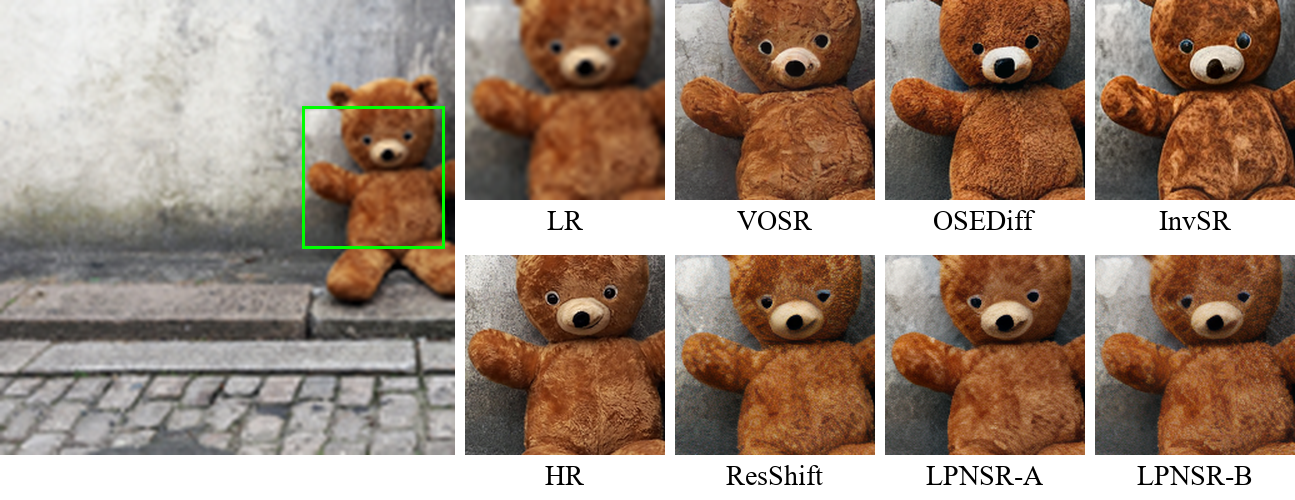}
\end{subfigure}
\caption{Additional qualitative comparison with other methods. (Zoom in for best view)}
\label{fig:more_qualitative_2}
\end{figure}

We provide additional qualitative comparisons in Figs.~\ref{fig:model_comparison_3} and~\ref{fig:more_qualitative_2}.

\end{document}